\documentclass[11pt,a4paper]{llncs}
\usepackage{amsmath}
\usepackage{amssymb}
\usepackage{graphicx}
\usepackage{marvosym}
\usepackage{url}
\usepackage{fancyhdr}

\usepackage{geometry}
\usepackage{subfig}

\newcommand{\keywords}[1]{\par\addvspace\baselineskip
\noindent\keywordname\enspace\ignorespaces#1}

\fancypagestyle{firstpage}{\fancyhf{}
}

\begin{document}


\title{\LARGE{Direct Punjabi to English speech translation using discrete units}}


%
%
\author{\large{Prabhjot Kaur\textsuperscript{1}  \and L. Andrew M. Bush\textsuperscript{2} \and Weisong Shi\textsuperscript{3}}}
\institute{\large{\textsuperscript{1}Wayne State University\\ \textsuperscript{2}Utah State University\\ \textsuperscript{3}University of Delaware}}

%


%
%


\maketitle

\thispagestyle{firstpage}

\begin{abstract}
Speech-to-speech translation is yet to reach the same level of coverage as text-to-text translation systems. The current speech technology is highly limited in its coverage of over 7000 languages spoken worldwide, leaving more than half of the population deprived of such technology and shared experiences. With voice-assisted technology (such as social robots and speech-to-text apps) and auditory content (such as podcasts and lectures) on the rise, ensuring that the technology is available for all is more important than ever. Speech translation can play a vital role in mitigating technological disparity and creating a more inclusive society. With a motive to contribute towards speech translation research for low-resource languages, our work presents a direct speech-to-speech translation model for one of the Indic languages called Punjabi to English. Additionally, we explore the performance of using a discrete representation of speech called discrete acoustic units as input to the Transformer-based translation model. The model, abbreviated as Unit-to-Unit Translation (U2UT), takes a sequence of discrete units of the source language (the language being translated from) and outputs a sequence of discrete units of the target language (the language being translated to). Our results show that the U2UT model performs better than the Speech-to-Unit Translation (S2UT) model by a 3.69 BLEU score.    
\end{abstract}


\section{Introduction}

Speech technology can play a vital role in bridging the gap between cultures of the world, fostering the exchange of ideas, and enabling more shared experiences. However, the current state of the speech technology is far from being inclusive to over 7000 languages worldwide\footnote{\url{https://www.ethnologue.com/insights/how-many-languages/}} \cite{black2019cmu}, \cite{tan2021survey}, \cite{pratap2023scaling}. Several automated services offer speech translation, including industry leaders like Google Translate, DeepL, and Bing Microsoft Translator. However, most of these systems rely on a 2-stage or 3-stage cascaded approach to speech translation. For example, a 3-stage cascaded approach involves integrating Automatic Speech Recognition (ASR), Machine Translation (MT), and Text-to-Speech (TTS) at its core. This design limits the availability of these text-based speech translation services to languages for which these three underlying subsystems are available, resulting in a notable absence of support for low-resource languages and those that are without a written form\footnote{\url{https://speechbot.github.io/}} \cite{chen2022speech}, \cite{barrault2023seamlessm4t}. Additionally, languages with accent types, tonal languages, and those with special sounds pose challenges to text-based translations. Therefore, the direct Speech-to-Speech Translation (S2ST) approach is a more practical and inclusive alternative to traditional cascade approaches. Though direct S2ST methods are still in their infancy and currently lag in performance compared to their cascaded counterparts, they promise the benefit of lower computational cost and inference latency. They are inherently less prone to error propagation\cite{lee2021direct}, \cite{barrault2023seamlessm4t} and have other benefits, such as preserving prosodic features, thus generating a more natural translation \cite{kharitonov2022textless}.

The direct S2ST is an end-to-end approach that ``directly'' converts speech in the source language to speech in the target language without going through a pipeline of ASR, MT, and TTS models. In contrast, the traditional cascade approach relies on intermediate text translation as it goes through a sequence of steps; namely, the speech in the source language is first converted to text in the source language by an ASR model, followed by source text to target text translation using the MT model. Finally, the translated text is processed through a TTS model that synthesizes target speech from the target text. Such a cascaded approach suffers from error propagation and accumulation from downstream ASR, MT, and TTS models, in addition to other limitations mentioned earlier. Our paper presents a method for direct S2ST called Unit-to-Unit Translation (U2UT). We demonstrate the approach through a Punjabi to English language pair, though the method works for any other pair of languages. Punjabi is the world's 36th most-spoken language, with over 51.7 million speakers.\footnote{\url{https://www.ethnologue.com/insights/ethnologue200/}} The limited research in NLP for the Punjabi language can be attributed to limited data availability in the digital form to train NLP models for various tasks \cite{gala2023indictrans2}, \cite{kaur2021automatic}, \cite{premjith2019neural} and lack of global benchmarks \cite{song2023globalbench}. Researchers often collect and use their own data in such cases, making it challenging to compare results \cite{premjith2019neural}, \cite{dua2022developing}. 

\subsection{Motivation}
\label{motivation}
Our research specifically sought to i) explore the performance of discretized speech representation approaches and ii) develop speech translation for low resource languages. The raw/continuous speech signals are typically sampled between 16kHz - 44kHz for downstream speech processing tasks. The sampled signals are then either used as such \cite{kaur2022fall} or are converted to a frequency representation of a signal (spectrogram, MFCC) \cite{jia2019direct}, \cite{lee2021direct} for feeding as input to the machine learning models. The sampled speech and its frequency representation are high dimensional and have redundancies \cite{chang2023exploring}. A recent development in speech processing is the use of discrete representations of speech called acoustic discrete units (also referred to as discrete speech units or simply as discrete units). These speech units are learned from a large speech corpus in a self-supervised learning fashion \cite{baevski2020wav2vec}, \cite{hsu2021hubert}. The benefit of using a discrete representation of speech is that it significantly reduces the dimension of the speech signal while still preserving the original speech content. \cite{lee2021direct} was the first work to utilize acoustic discrete units for speech-to-speech translation tasks. However, it utilizes discrete units only for the target speech, while the source speech still uses frequency representation. Motivated by the recent performance gains of using acoustic discrete units in various speech processing tasks \cite{chou2023toward}, \cite{chang2023exploring}, our goal is to study the impacts of using acoustic discrete units for both source and target speech in the speech-to-speech translation task. 

\subsection{Contribution}
\label{Contribution}
The contribution of our work is as follows. Firstly, we introduce a Transformer-based, Unit-to-Unit Translation (U2UT) model for direct speech translation. Unlike prior approaches that rely on spectrograms or raw audio as input, our model leverages discrete acoustic units to represent source speech. Since speech is represented in a lower dimension compared to the raw audio or spectrograms, discrete unit representation has the advantage of lower memory footprint and computational cost. The exploration of discrete acoustic units to represent input speech is one of the contributions of our work. Our findings demonstrate that the U2UT model achieves a BLEU score that is 3.69 points higher than a prior state-of-the-art method which uses a spectrogram to represent input speech. It leads us to conclude that acoustic discrete units are sufficient to represent speech and offer a superior representation compared to spectrograms. The second contribution of our research is to advance the speech translation research by focusing on low-resource languages. Through our work, we release a direct speech-to-speech translation model for Punjabi to English. 

\begin{figure}[h!]
    \centering
    \includegraphics[width=1.00\linewidth]{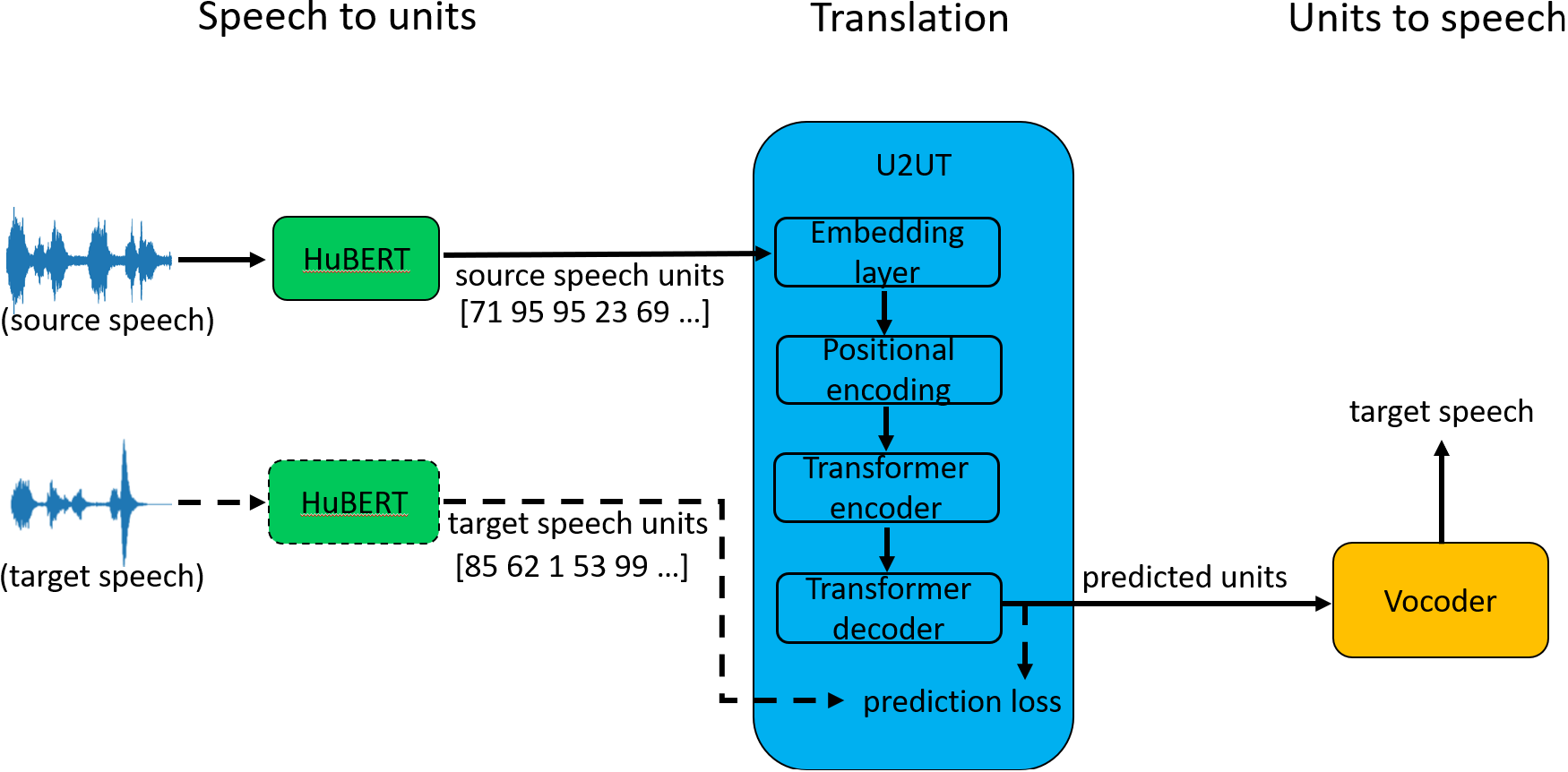}
    \caption{The overall framework of direct speech-to-speech translation using acoustic discrete units: It consists of (1) a Pre-processing step that converts speech to units, (2) a Transformer based Unit to Unit Translation (U2UT) model, (3) a Post-processing step that converts predicted target units to target speech. The dotted lines represent part of the framework that runs only during model training.}
    \label{model_architecture}
\end{figure}
\hfill 

The rest of the paper is organized as follows. Section \ref{RelatedWork} reviews existing direct S2ST models and parallel speech corpora, followed by Section \ref{methodology}, which describes the dataset used in our work and the model framework in detail. Section \ref{results} discusses the results and experiments conducted. Section \ref{limitations} highlights the limitations of our approach and future directions, followed by a conclusion in Section \ref{conclusion}.

\section{Related Work}
\label{RelatedWork}
\subsection{Direct Speech-to-Speech Translation (S2ST) models}
Direct S2ST is a recent development in speech translation research. Within the domain, there are two approaches to translation when given the source speech. Both approaches output speech representations. In the first approach, the translation model is trained to predict the spectrogram for the target speech. In the second and newer approach, the translation model predicts discrete acoustic units \cite{hsu2021hubert} for target speech. These speech representations from the translation model are converted to a raw audio waveform using a vocoder as the final output. \cite{lee2021direct} demonstrates that the second approach yields better results. 

Translatotron \cite{jia2019direct} is the first direct S2ST model. The model consists of an attention-based encoder-decoder architecture that converts a spectrogram for a speech in the source language to a spectrogram for a speech in the target language, a vocoder that converts the generated spectrogram to a waveform, and finally, a speaker encoder that allows the model to preserve the voice of the source speaker throughout the translation process. The encoder and decoder both use layers of LSTMs. Additionally, the authors emphasize that multitask training is essential, which in this case is integrated by including decoders for the auxiliary tasks of predicting the phonemes of source and target languages. The model is tested on two different datasets for English and Spanish. The results show that the performance of Translatotron is comparable to the conventional cascade approaches, with Translatotron slightly underperforming the conventional approaches.
Further, \cite{jia2022translatotron} improves upon the results of the Translatotron model and presents a Translatotron 2 model. The architecture consists of an encoder, a decoder, and a synthesizer. All these components are connected through a single attention module. Translatotron 2 uses conformer \cite{gulati2020conformer} as an encoder, LSTMs as a decoder, and multi-head attention for attention. The paper states that the overall architecture leads to a performance boost in the model rather than the individual sub-components. Both Translatotron 1 and 2 predict the spectrogram of the target speech given the source speech. 

Next, \cite{lee2021direct} presents a first Speech-to-Unit Translation (S2UT) model, which predicts discrete units for the target speech instead of predicting the spectrograms. It is a Transformer\cite{vaswani2017attention} based model. It takes the Mel-Frequency Cepstral Coefficient (MFCC) features derived from the source language speech as input to the encoder layer, followed by multi-head attention. The decoder then predicts a sequence of discrete units for the target language from the encoded input. It is important to note that the overall model is trained using discrete units obtained from a pre-trained HuBERT model \cite{hsu2021hubert}. Finally, a separately trained vocoder converts the discrete units to a raw audio waveform (speech). The model exploits multitask learning and pre-training. For multitask learning, the model uses auxiliary tasks such as predicting the phonemes, characters, and text of the target or the source language. The model is tested on the Fisher dataset for Spanish to English translation tasks. Further, \cite{lee2021textless} extends the S2UT model in \cite{lee2021direct} to training and testing with the real-world S2ST data in multiple languages. Specifically, the authors propose a novel speech normalization technique that minimizes the variation in speech from multiple speakers and recording environments. The speech normalizer is created by fine-tuning the HuBERT \cite{hsu2021hubert} model with a speech from a reference speaker. This speech normalizer is then used to create targets (labels) for training the S2UT model \cite{lee2021direct}. The experiments on the VoxPopuli dataset show that speech normalization is essential in increasing the effectiveness of the S2UT model. \cite{popuri2022enhanced} provide further improvements to S2UT model by incorporating pre-training. \cite{gong2023multilingual} extends S2UT \cite{lee2021direct} to from bilingual to multilingual. It supports S2ST from English to sixteen other target languages. Finally, SeamlessM4T \cite{barrault2023seamlessm4t} is the latest model in the S2ST domain. It is a multitask, multimodal, and multilingual model that supports speech-to-speech translation and other tasks such as speech-to-text translation and text-to-speech translation. The language coverage for SeamlessM4T is available here\footnote{\url{https://github.com/facebookresearch/seamless_communication/blob/main/docs/m4t/README.md}}. It uses UnitY model \cite{inaguma2022unity} at its core and predicts target discrete units. Other notable works in this area include Transformer based direct speech-to-speech translation with transcoder \cite{kano2021transformer}, AudioPaLM \cite{rubenstein2023audiopalm}, direct translation between Hokkien and English language pairs \cite{chen2022speech}.

It is important to note that S2ST (either using a direct or cascade approach) for Indian languages is still an under-explored area \cite{mujadia-sharma-2023-towards}. It is especially true for the Punjabi language. \cite{9612485} and \cite{mujadia-sharma-2023-towards} are the only pieces of work we have found that develop English-to-Punjabi speech translation pipelines. Both of these works use a cascade approach. As highlighted earlier, building a pipeline for English to Punjabi speech translation using the cascade approach is limited by the unavailability of state-of-the-art ASR \cite{singh2020asroil}, \cite{kaur2021automatic}, MT \cite{jha2022review}, and S2T models for Punjabi. Specifically, there is no S2T model available for Punjabi \cite{9612485}. Luckily, there is a growing interest from the research community to develop speech technologies for all languages worldwide. Some of the works include \cite{doddapaneni2023towards}, \cite{song2023globalbench}, No Language Left Behind \cite{costa2022no}, and Massively Multilingual Speech \cite{pratap2023scaling}. These models are incrementally supporting additional languages. For example, the recent release of the seamlessM4T model \cite{barrault2023seamlessm4t} supports direct S2ST for Punjabi to English direction. However, it does not support English to Punjabi speech translation. Therefore, more work is needed to develop speech translation technologies for lower-resource languages like Punjabi.

\subsection{S2ST dataset}
The direct S2ST models require parallel speech data in source and target languages for training. The scarcity of parallel speech data, especially for Punjabi and English pairs, is critical. Table \ref{tab1} provides a comprehensive overview of the currently available multilingual datasets. The table further highlights the datasets that include parallel speech data for Punjabi and English (the last column), accentuating our focus's urgency and significance.

It is important to note that there are other multilingual speech datasets available publicly not included in Table \ref{tab1}, such as Multilinguagl LibriSpeech dataset \cite{pratap2020mls} supporting 8 languages and Common Voice \cite{ardila2019common} containing over 70 languages, and CoVOST 2 \cite{wang2020covost} derived from the Common Voice corpus (Check again if CoVOST 2 contains parallel speech to speech dataset), BABEL speech corpus\cite{gales2014speech}, Voxlingua107 \cite{valk2021voxlingua107}, Indian languages datasets from AI4Bharat\footnote{\url{https://ai4bharat.iitm.ac.in/}.}. However, these datasets do not provide parallel speech in a given pair of languages; hence, they are not included in the table.

\begin{table*}[htbp]
\centering
\begin{tabular}{|p{3.00cm}|p{5.25cm}|p{1.75cm}|p{1.75cm}|p{2.75cm}|}
\hline
\textbf{Dataset}& \textbf{Description} & \textbf{Parallel text} & \textbf{Parallel speech} &\textbf{Parallel speech for English and Punjabi}\\
\hline
CVSS \cite{jia2022cvss} & sentence level speech to speech translation pairs from 21 other languages to English, derived from Common Voice and CoVoST2 & Yes & Yes&No\\
\hline
VoxPopuli \cite{wang2021voxpopuli} & spontaneous speech, parallel speech to speech translation dataset for 15 languages  & Yes &Yes &No\\
\hline
CMU Wilderness \cite{black2019cmu} & read speech, parallel speech to speech translation dataset for 700 languages  & Yes &Yes &No\\
\hline
MaSS \cite{boito2019mass}& read speech, derived from CMU Wilderness dataset \cite{black2019cmu}, sentence level speech to speech parallel dataset for 8 language pairs & Yes & Yes&No\\
\hline
MuST-C \cite{di2019must} & spontaneous speech, English to 8 other languages translation of the TED talks &Yes  & No &No\\
\hline
SpeechMatrix \cite{duquenne2022speechmatrix} & spontaneous speech, 136 language pairs translated from European Parliament recordings  &Yes  & Yes &No\\
\hline
\textbf{Fleurs} \cite{conneau2022fleurs} & read speech, parallel speech data for 102 languages & Yes & Yes & \textbf{Yes} \\
\hline
\end{tabular}
\caption{Summary of public multilingual datasets.}
\label{tab1}
\end{table*}

\section{Methodology}
\label{methodology}
As mentioned in Section \ref{RelatedWork}, the current S2ST models predict either the spectrogram (as in Translatotron\cite{jia2019direct}) or the discrete units (as in S2UT\cite{lee2021direct}) of the target language. However, both model types use spectrogram or its derivative, such as MFCC of the source speech as an input. Our model takes discrete units of the source language as an input and predicts discrete units of the target language as an output, thus the name U2UT. Section \ref{dataset} explains the dataset used for training and testing the model, followed by Section \ref{model}, which provides model details, training, and evaluation framework.
\subsection{Data}
\label{dataset}
FLEURS dataset \cite{conneau2022fleurs} is the only publicly available data with parallel speech for English and Punjabi. However, with only 1625 Punjabi and English pairs in the training set, it is too small to train an S2ST model from scratch. Therefore, we first prepare a parallel speech dataset for English and Punjabi using a large-scale ASR dataset called Kathbath \cite{javed2022indicsuperb}.
Kathbath\footnote{\url{https://github.com/AI4Bharat/indicSUPERB}.} is an ASR dataset that contains read speech and transcript pairs for 12 Indian languages, including Punjabi. It contains around 136 hours of Punjabi speech from several native speakers. The average sentence length of samples in the dataset is between 5-10 words. The dataset has train, dev, and test subsets containing 83578 samples, 3270 samples, and 3202 samples. Since we need parallel speech data for training a direct S2ST model for English and Punjabi, we first create a parallel English speech and text from the Punjabi subset of the Kathbath dataset. We accomplish this using the SeamlessM4T\footnote{\url{https://github.com/facebookresearch/seamless_communication}.} model. The translated English speech and text generated by SeamlessM4T are manually verified for quality by a proficient speaker in Punjabi and English. It is important to note that while Punjabi speech is natural, English speech is all synthetic generated by the SeamlessM4T model. Finally, both English and Punjabi audios have a 16kHz sampling rate. 

\subsection{Model framework and training}
\label{model}
We implement the Transformer \cite{vaswani2017attention} model and train it to predict the target language's discrete units. The discrete units of the source language are given as input to the Transformer encoder. The Transformer decoder takes the encoded representation of the discrete units of the source language to produce a sequence of discrete units of the target language. The training is done in a teacher-forcing manner, using discrete units of the actual/true output sequence to learn to generate the next actual unit. Therefore, we must first create discrete units for both source and target languages to train the model. 

\subsubsection{Convert speech to units}
\label{HuBERT}
\begin{figure}%
    \centering
    \subfloat[\centering Punjabi subset]{{\includegraphics[width=7 cm]{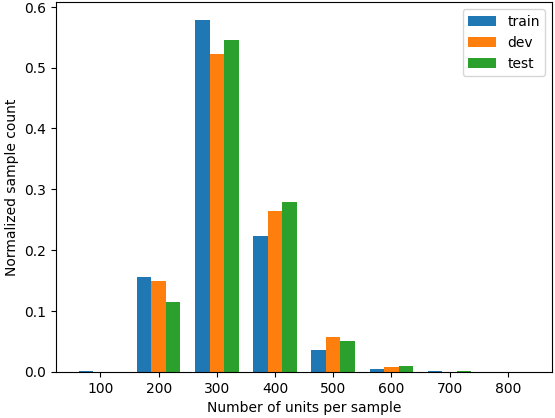} }}%
    \qquad
    \subfloat[\centering English subset]{{\includegraphics[width=7 cm]{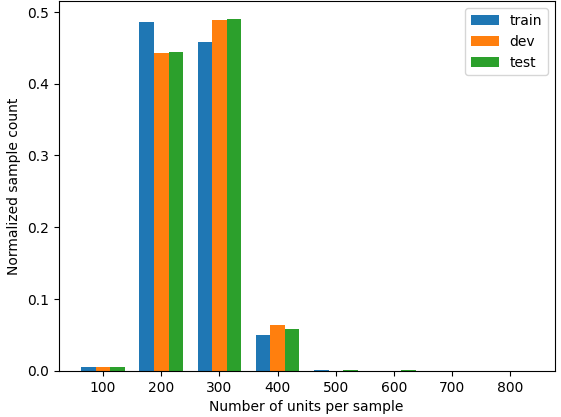} }}%
    \caption{Sample count vs number of discrete units per sample in English and Punjabi subsets of Kathbath dataset}%
    \label{Units}%
\end{figure}

The discrete units for both target and source languages are derived using a pre-trained HuBERT Base model\footnote{\url{https://github.com/facebookresearch/fairseq/blob/main/examples/hubert/README.md}} \cite{hsu2021hubert}, trained on 960 hours of Librispeech \cite{panayotov2015librispeech} corpus. Similar to \cite{lee2021direct}, we use the 6th layer of the HuBERT model and derive the discrete units using 100 clusters. The discrete units are created from raw English and Punjabi audio. The average number of discrete units in a given sentence is about 300, as shown in Figure \ref{Units}.    Finally, we create pairs of sequences of discrete units for the source and the target languages to train the model. Further, since the original audio samples are of different lengths, the resulting sequence of discrete units is also of varying lengths for different samples in the dataset. We pad the shorter sequences with 0s and trim the longer sequences to ensure that all of them are of sequence length of 300 before they are fed into the model for training.

\subsubsection{Unit-to-Unit Translation (U2UT)}
\label{U2UT}
Next, we build a sequence-to-sequence model based on the Transformer architecture that learns to translate the source language's discrete units to the target language's discrete units. The best model contains three encoder layers, three decoder layers, and one head, and the embedding size (d\_model) is set to 512. The model is trained with a learning rate of 0.0001, Adam optimizer, 0.1 dropout rate, 25 batch size, and cross-entropy loss. After training for 80 epochs, we save the model for evaluation and inference.

\subsubsection{Convert predicted units to target speech}
\label{vocoder}
Since the end goal is to obtain the audio in the target language, an additional step is required to convert this predicted sequence of discrete units to a waveform. A Vocoder does the speech synthesis. We use a pre-trained Vocoder called unit-based HiFi-GAN\footnote{\url{https://github.com/facebookresearch/fairseq/blob/main/examples/speech_to_speech/docs/direct_s2st_discrete_units.md}} from \cite{lee2021direct} for this step. The overall framework of our approach consisting of all three steps mentioned above is shown in Figure  \ref{model_architecture}.

\subsection{Model evaluation}
We evaluate the trained model on the test set consisting of 3202 samples. First, we derive discrete units of the source language using the HuBERT model as described in Section \ref{HuBERT}. The source discrete units are fed to the model to predict target discrete units in a greedy fashion. The target discrete units are then converted to raw speech using the Vocoder as a post-processing step. Eventually, we have a target speech. To quantify how much the predicted target speech matches the ground truth, we convert the translated audio to text using an ASR model (specifically the Whisper large-v3\footnote{\url{https://huggingface.co/openai/whisper-large-v3}} \cite{radford2022robust}) and compare the Whisper generated transcripts to ground truth text for the target language. The Whisper model is chosen during evaluation as it is the state-of-the-art model for ASR. It is a multitask speech recognition model trained on 680,000 hours of multilingual speech. We report the final result of our method as a BiLingual Evaluation Understudy (BLEU) score \cite{papineni2002bleu}. We use the SacreBLEU \cite{post2018call} method.

\section{Results and discussion}
\label{results}
To compare the results of our method with previous work published in speech translation research, we compare the results with both cascaded and direct speech-to-speech translation approaches. For the direct approach, we compare with the S2UT model \cite{lee2021direct} as it also involves predicting discrete units of the target language. We begin with training the original S2UT model available in the fairseq\footnote{\url{https://github.com/facebookresearch/fairseq/tree/main/examples/speech_to_speech}} library on the same dataset as was used to train our U2UT model. The S2UT model takes the spectrograms of the source speech as an input and is trained to predict the sequence of discrete units of the target speech. Similar to our work, the S2UT model also involves a data pre-processing step to create discrete units of target language using the HuBERT model for training. However, it needs the discrete units only for the target speech. We trained with multitasking (joint text and speech) and without multitasking and noticed no significant difference in the performance. The model is trained for 50 epochs. The model architecture contains six encoder layers, six decoder layers, four heads, 256 dimensions of the encoder embedding, and 2048 dimensions for the feed-forward. The sequence of target discrete units predicted by the trained model is converted to speech using the Vocoder. Finally, we use the Whisper ASR model to transcribe the predicted speech and report the BLEU score.

Next, to compare our results with a traditional cascaded approach, we choose Whisper. It uses a 2-stage approach (ASR followed by MT) for speech-to-text translation. The Whisper ASR first converts Punjabi speech input to Punjabi text, and then the Whisper MT translates Punjabi text to English text. The generated English text is compared with the ground truth transcripts. The results of our method, S2UT, and Whisper are shown in Table \ref{tab2}. We also show samples of generated translations by all three models in Table \ref{tab3}. Our model attempts to reproduce the sounds in the original speech but does not quite output the original word. Our method outperforms the S2UT model. We conclude that using acoustic discrete units are sufficient to represent input speech compared to the spectrogram representation. We would like to emphasize that the S2UT model used for comparing our results was used as a black-box model, meaning we used the default model parameters for training. Further, both approaches lag in performance compared to the cascaded approach. We believe that the significant difference in the performance is due to the sample size used for training. U2UT and S2UT models were trained with 83578 samples, whereas the Whisper model was trained on a much larger dataset. Access to large-scale parallel speech data is challenging for building direct speech-to-speech translation systems that outperform cascaded approaches.

\begin{table*}[h]
\centering
\begin{tabular}{|p{2.75cm}|p{6.75cm}|p{2.25cm}|}

\hline
\textbf{Method}& \textbf{Description} & \textbf{BLEU ($\uparrow$)}\\
\hline
U2UT (\textbf{Ours})& source discrete in and target discrete out& 3.9829   \\
\hline
S2UT \cite{lee2021direct}& source spectrogram in and target discrete out&  0.2954 
\\
\hline
2-stage cascade approach (Whisper) \cite{radford2022robust} & ASR followed by MT & 18.1136 
\\
\hline
\end{tabular}

\caption{Results for Punjabi to English translation using Kathbath test set.}
\label{tab2}
\end{table*}

\begin{table*}[h]
\centering
\begin{tabular}{|p{2.50cm}|p{3.00cm}|p{3.00cm}|p{3.00cm}|p{3.00cm}|}
\hline
\textbf{Audio sample ID} (Kathbath test set) &\textbf{Actual transcript} & \textbf{Prediction (U2UT)} & \textbf{Prediction (Whisper)} & \textbf{Prediction (S2UT)} \\
\hline
844424933626045-586-m.wav& doctor gurbaksh singh frank explains the culture in this way&actor gerbache saying fraud & doctor gurubaksh singh frank describes the practice in this way  & this is necessary to talk about the process of the country\\
\hline
844424933639213-32-m.wav&he also wrote sketches memoirs biographies travelogues and translated some works&he is wrecked at measure the post of the resident of the cagular and some novel & he also translated the book of kujra chanaama in rekha chitra yadda jeevaniyaan falsafah safar naam & the ceremony was given by the punjab government by the chief minister of punjab \\
\hline
844424933647726-601-m.wav&the greed of the ministers to come home with money is turned upside down &determined by finnish minister is provided by the people &  minister says that he will give money to the greedy people & it is very difficult to imagination and investigation of the country\\
\hline
844424931569220-989-f.wav& congress workers started distributing cheap grain at dharmakot bagga& the congress said the pack and congress was began to started& the term court has declared that the  & it is very difficult to imagination in the country\\
\hline
844424933162326-703-f.wav& n r i ravindra kakku inaugurated the school s kitchens& dees views officers of the schools of faith prona cases officerer& i am ravindra kakku from the school of dehep kaaleya uttakadam  & it is very difficult to imagination in the country \\
\hline
\end{tabular}

\caption{Sample output (Transcripts corresponding to the audio translated from Punjabi to English by various models).}
\label{tab3}
\end{table*}

\begin{table*}[h!]
\centering
\begin{tabular}{|p{1.00cm} |p{1.50cm}|p{1.00cm}| p{1.75cm}| p{2.00cm} | p{1.75cm}| p{1.25cm}| p{1.25cm}| p{1.25cm}| p{1.25cm} |}
\hline
\textbf{Expt. No.}&\textbf{Sequence length} & \textbf{Heads} & \textbf{Encoder-Decoder layers} & \textbf{Feedforward dimension} & \textbf{Learning rate}&\textbf{Epochs} & \textbf{Val loss} & \textbf{BLEU} ($\uparrow$)& \textbf{WER} ($\downarrow$)\\
\hline
1&300&1 & 3 &2048& 0.0001&80 & 0.006462& 3.9829&99.62\\
\hline
2&300&1 & 3 &2048& 0.00001&100 & 0.02434& 1.0601 &101.52\\
\hline
3&300&1  & 3 & 1024&0.0001&100& 0.01&3.4736 &99.66\\
\hline
4&300&1 & 6 & 1024&0.0001&80 &0.004758 & 3.3838&100.50\\
\hline
5&200&1 & 3 &1024& 0.0001&80 & 0.010644& 2.4202&99.77\\
\hline
6&300&4 & 6 &1024&0.0001&100 & 0.001998& 1.3939&107.02\\
\hline
7&300&1 & 3 &2048& 0.001&100 & 0.021073&0.1922 &103.83\\
\hline
\end{tabular}
\caption{Ablation study: Results for various model parameters.}
\label{tab4}
\end{table*}

\begin{figure}[h]
    \centering
    \includegraphics[width=0.65\linewidth]{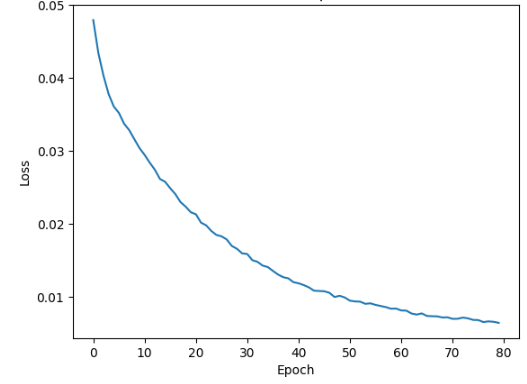}
    \caption{Validation loss for Exp1. 1 in Table \ref{tab4}.}
    \label{val_loss}
\end{figure}

\section{Limitations and future directions}
\label{limitations}
The limitations of our study were the dataset size and available computing. We had 83578 samples for training, which is much smaller than the dataset used in the state-of-the-art models in speech processing, such as Whisper and SeamlessM4T. Further, the limitation of using acoustic discrete units for source speech as input to the machine learning model is that it introduces an additional pre-processing step of obtaining the discrete units. A possible future direction could be to leverage transfer learning rather than training the model from scratch to mitigate data scarcity issues. There is also a need for more parallel speech datasets for Punjabi and English pairs.

\section{Conclusion}
\label{conclusion}
We have created a direct speech-to-speech translation model called U2UT. Specifically, we investigated using discrete representations of speech, called acoustic discrete units, to represent input speech in the direct speech-to-speech translation model. Previous work in direct speech translation research uses frequency representations such as spectrograms and MFCC to represent input speech. Spectrograms are high-dimensional and consume more memory for storage and processing. Discrete unit representation offers an alternative that is much lower in dimension. Our results show that using discrete units to represent input speech is a promising future direction for direct speech-to-speech translation. We achieved a BLEU score of 3.9829, which is 3.69 points higher than the BLEU score achieved by an S2UT model that uses spectrograms to represent input speech. We demonstrated the performance of the U2UT model for Punjabi to English translation, although the method can be applied to any two pairs of languages. Additionally, the choice of the Punjabi language is deliberate, as it is one of the low-resource languages. We need to focus research efforts on such languages, as this is a step to ensure that speech technology is available to people who speak languages other than English. Lastly, as a future direction toward improving the performance of our translation system, we will focus on transfer learning and collecting more parallel speech data.


\bibliography{bibtex/bib/IEEEabrv.bib,bibtex/bib/IEEEexample.bib}{}

\begin{thebibliography}{10}
\providecommand{\url}[1]{#1}
\csname url@samestyle\endcsname
\providecommand{\newblock}{\relax}
\providecommand{\bibinfo}[2]{#2}
\providecommand{\BIBentrySTDinterwordspacing}{\spaceskip=0pt\relax}
\providecommand{\BIBentryALTinterwordstretchfactor}{4}
\providecommand{\BIBentryALTinterwordspacing}{\spaceskip=\fontdimen2\font plus
\BIBentryALTinterwordstretchfactor\fontdimen3\font minus \fontdimen4\font\relax}
\providecommand{\BIBforeignlanguage}[2]{{%
\expandafter\ifx\csname l@#1\endcsname\relax
\typeout{** WARNING: IEEEtran.bst: No hyphenation pattern has been}%
\typeout{** loaded for the language `#1'. Using the pattern for}%
\typeout{** the default language instead.}%
\else
\language=\csname l@#1\endcsname
\fi
#2}}
\providecommand{\BIBdecl}{\relax}
\BIBdecl

\bibitem{black2019cmu}
A.~W. Black, ``Cmu wilderness multilingual speech dataset,'' in \emph{ICASSP 2019-2019 IEEE International Conference on Acoustics, Speech and Signal Processing (ICASSP)}.\hskip 1em plus 0.5em minus 0.4em\relax IEEE, 2019, pp. 5971--5975.

\bibitem{tan2021survey}
X.~Tan, T.~Qin, F.~Soong, and T.-Y. Liu, ``A survey on neural speech synthesis,'' \emph{arXiv preprint arXiv:2106.15561}, 2021.

\bibitem{pratap2023scaling}
V.~Pratap, A.~Tjandra, B.~Shi, P.~Tomasello, A.~Babu, S.~Kundu, A.~Elkahky, Z.~Ni, A.~Vyas, M.~Fazel-Zarandi \emph{et~al.}, ``Scaling speech technology to 1,000+ languages,'' \emph{arXiv preprint arXiv:2305.13516}, 2023.

\bibitem{chen2022speech}
P.-J. Chen, K.~Tran, Y.~Yang, J.~Du, J.~Kao, Y.-A. Chung, P.~Tomasello, P.-A. Duquenne, H.~Schwenk, H.~Gong \emph{et~al.}, ``Speech-to-speech translation for a real-world unwritten language,'' \emph{arXiv preprint arXiv:2211.06474}, 2022.

\bibitem{barrault2023seamlessm4t}
L.~Barrault, Y.-A. Chung, M.~C. Meglioli, D.~Dale, N.~Dong, P.-A. Duquenne, H.~Elsahar, H.~Gong, K.~Heffernan, J.~Hoffman \emph{et~al.}, ``Seamlessm4t-massively multilingual \& multimodal machine translation,'' \emph{arXiv preprint arXiv:2308.11596}, 2023.

\bibitem{lee2021direct}
A.~Lee, P.-J. Chen, C.~Wang, J.~Gu, S.~Popuri, X.~Ma, A.~Polyak, Y.~Adi, Q.~He, Y.~Tang \emph{et~al.}, ``Direct speech-to-speech translation with discrete units,'' \emph{arXiv preprint arXiv:2107.05604}, 2021.

\bibitem{kharitonov2022textless}
E.~Kharitonov, J.~Copet, K.~Lakhotia, T.~A. Nguyen, P.~Tomasello, A.~Lee, A.~Elkahky, W.-N. Hsu, A.~Mohamed, E.~Dupoux \emph{et~al.}, ``textless-lib: A library for textless spoken language processing,'' \emph{arXiv preprint arXiv:2202.07359}, 2022.

\bibitem{gala2023indictrans2}
J.~Gala, P.~A. Chitale, R.~AK, S.~Doddapaneni, V.~Gumma, A.~Kumar, J.~Nawale, A.~Sujatha, R.~Puduppully, V.~Raghavan \emph{et~al.}, ``Indictrans2: Towards high-quality and accessible machine translation models for all 22 scheduled indian languages,'' \emph{arXiv preprint arXiv:2305.16307}, 2023.

\bibitem{kaur2021automatic}
J.~Kaur, A.~Singh, and V.~Kadyan, ``Automatic speech recognition system for tonal languages: State-of-the-art survey,'' \emph{Archives of Computational Methods in Engineering}, vol.~28, pp. 1039--1068, 2021.

\bibitem{premjith2019neural}
B.~Premjith, M.~A. Kumar, and K.~Soman, ``Neural machine translation system for english to indian language translation using mtil parallel corpus,'' \emph{Journal of Intelligent Systems}, vol.~28, no.~3, pp. 387--398, 2019.

\bibitem{song2023globalbench}
Y.~Song, C.~Cui, S.~Khanuja, P.~Liu, F.~Faisal, A.~Ostapenko, G.~I. Winata, A.~F. Aji, S.~Cahyawijaya, Y.~Tsvetkov \emph{et~al.}, ``Globalbench: A benchmark for global progress in natural language processing,'' \emph{arXiv preprint arXiv:2305.14716}, 2023.

\bibitem{dua2022developing}
S.~Dua, S.~S. Kumar, Y.~Albagory, R.~Ramalingam, A.~Dumka, R.~Singh, M.~Rashid, A.~Gehlot, S.~S. Alshamrani, and A.~S. AlGhamdi, ``Developing a speech recognition system for recognizing tonal speech signals using a convolutional neural network,'' \emph{Applied Sciences}, vol.~12, no.~12, p. 6223, 2022.

\bibitem{kaur2022fall}
P.~Kaur, Q.~Wang, and W.~Shi, ``Fall detection from audios with audio transformers,'' \emph{Smart Health}, vol.~26, p. 100340, 2022.

\bibitem{jia2019direct}
Y.~Jia, R.~J. Weiss, F.~Biadsy, W.~Macherey, M.~Johnson, Z.~Chen, and Y.~Wu, ``Direct speech-to-speech translation with a sequence-to-sequence model,'' \emph{arXiv preprint arXiv:1904.06037}, 2019.

\bibitem{chang2023exploring}
X.~Chang, B.~Yan, K.~Choi, J.~Jung, Y.~Lu, S.~Maiti, R.~Sharma, J.~Shi, J.~Tian, S.~Watanabe \emph{et~al.}, ``Exploring speech recognition, translation, and understanding with discrete speech units: A comparative study,'' \emph{arXiv preprint arXiv:2309.15800}, 2023.

\bibitem{baevski2020wav2vec}
A.~Baevski, Y.~Zhou, A.~Mohamed, and M.~Auli, ``wav2vec 2.0: A framework for self-supervised learning of speech representations,'' \emph{Advances in neural information processing systems}, vol.~33, pp. 12\,449--12\,460, 2020.

\bibitem{hsu2021hubert}
W.-N. Hsu, B.~Bolte, Y.-H.~H. Tsai, K.~Lakhotia, R.~Salakhutdinov, and A.~Mohamed, ``Hubert: Self-supervised speech representation learning by masked prediction of hidden units,'' \emph{IEEE/ACM Transactions on Audio, Speech, and Language Processing}, vol.~29, pp. 3451--3460, 2021.

\bibitem{chou2023toward}
J.-C. Chou, C.-M. Chien, W.-N. Hsu, K.~aLivescu, A.~Babu, A.~Conneau, A.~Baevski, and M.~Auli, ``Toward joint language modeling for speech units and text,'' \emph{arXiv preprint arXiv:2310.08715}, 2023.

\bibitem{jia2022translatotron}
Y.~Jia, M.~T. Ramanovich, T.~Remez, and R.~Pomerantz, ``Translatotron 2: High-quality direct speech-to-speech translation with voice preservation,'' in \emph{International Conference on Machine Learning}.\hskip 1em plus 0.5em minus 0.4em\relax PMLR, 2022, pp. 10\,120--10\,134.

\bibitem{gulati2020conformer}
A.~Gulati, J.~Qin, C.-C. Chiu, N.~Parmar, Y.~Zhang, J.~Yu, W.~Han, S.~Wang, Z.~Zhang, Y.~Wu \emph{et~al.}, ``Conformer: Convolution-augmented transformer for speech recognition,'' \emph{arXiv preprint arXiv:2005.08100}, 2020.

\bibitem{vaswani2017attention}
A.~Vaswani, N.~Shazeer, N.~Parmar, J.~Uszkoreit, L.~Jones, A.~N. Gomez, {\L}.~Kaiser, and I.~Polosukhin, ``Attention is all you need,'' \emph{Advances in neural information processing systems}, vol.~30, 2017.

\bibitem{lee2021textless}
A.~Lee, H.~Gong, P.-A. Duquenne, H.~Schwenk, P.-J. Chen, C.~Wang, S.~Popuri, J.~Pino, J.~Gu, and W.-N. Hsu, ``Textless speech-to-speech translation on real data,'' \emph{arXiv preprint arXiv:2112.08352}, 2021.

\bibitem{popuri2022enhanced}
S.~Popuri, P.-J. Chen, C.~Wang, J.~Pino, Y.~Adi, J.~Gu, W.-N. Hsu, and A.~Lee, ``Enhanced direct speech-to-speech translation using self-supervised pre-training and data augmentation,'' \emph{arXiv preprint arXiv:2204.02967}, 2022.

\bibitem{gong2023multilingual}
H.~Gong, N.~Dong, S.~Popuri, V.~Goswami, A.~Lee, and J.~Pino, ``Multilingual speech-to-speech translation into multiple target languages,'' \emph{arXiv preprint arXiv:2307.08655}, 2023.

\bibitem{inaguma2022unity}
H.~Inaguma, S.~Popuri, I.~Kulikov, P.-J. Chen, C.~Wang, Y.-A. Chung, Y.~Tang, A.~Lee, S.~Watanabe, and J.~Pino, ``Unity: Two-pass direct speech-to-speech translation with discrete units,'' \emph{arXiv preprint arXiv:2212.08055}, 2022.

\bibitem{kano2021transformer}
T.~Kano, S.~Sakti, and S.~Nakamura, ``Transformer-based direct speech-to-speech translation with transcoder,'' in \emph{2021 IEEE Spoken Language Technology Workshop (SLT)}.\hskip 1em plus 0.5em minus 0.4em\relax IEEE, 2021, pp. 958--965.

\bibitem{rubenstein2023audiopalm}
P.~K. Rubenstein, C.~Asawaroengchai, D.~D. Nguyen, A.~Bapna, Z.~Borsos, F.~d.~C. Quitry, P.~Chen, D.~E. Badawy, W.~Han, E.~Kharitonov \emph{et~al.}, ``Audiopalm: A large language model that can speak and listen,'' \emph{arXiv preprint arXiv:2306.12925}, 2023.

\bibitem{mujadia-sharma-2023-towards}
\BIBentryALTinterwordspacing
V.~Mujadia and D.~Sharma, ``Towards speech to speech machine translation focusing on {I}ndian languages,'' in \emph{Proceedings of the 17th Conference of the European Chapter of the Association for Computational Linguistics: System Demonstrations}.\hskip 1em plus 0.5em minus 0.4em\relax Dubrovnik, Croatia: Association for Computational Linguistics, May 2023, pp. 161--168. [Online]. Available: \url{https://aclanthology.org/2023.eacl-demo.19}
\BIBentrySTDinterwordspacing

\bibitem{9612485}
N.~Shaghaghi, S.~Ghosh, and R.~Kapoor, ``Classroute: An english to punjabi educational video translation pipeline for supporting punjabi mother-tongue education,'' in \emph{2021 IEEE Global Humanitarian Technology Conference (GHTC)}, 2021, pp. 342--348.

\bibitem{singh2020asroil}
A.~Singh, V.~Kadyan, M.~Kumar, and N.~Bassan, ``Asroil: a comprehensive survey for automatic speech recognition of indian languages,'' \emph{Artificial Intelligence Review}, vol.~53, pp. 3673--3704, 2020.

\bibitem{jha2022review}
A.~Jha and H.~Y. Patil, ``A review of machine transliteration, translation, evaluation metrics and datasets in indian languages,'' \emph{Multimedia Tools and Applications}, pp. 1--32, 2022.

\bibitem{doddapaneni2023towards}
S.~Doddapaneni, R.~Aralikatte, G.~Ramesh, S.~Goyal, M.~M. Khapra, A.~Kunchukuttan, and P.~Kumar, ``Towards leaving no indic language behind: Building monolingual corpora, benchmark and models for indic languages,'' in \emph{Proceedings of the 61st Annual Meeting of the Association for Computational Linguistics (Volume 1: Long Papers)}, 2023, pp. 12\,402--12\,426.

\bibitem{costa2022no}
M.~R. Costa-juss{\`a}, J.~Cross, O.~{\c{C}}elebi, M.~Elbayad, K.~Heafield, K.~Heffernan, E.~Kalbassi, J.~Lam, D.~Licht, J.~Maillard \emph{et~al.}, ``No language left behind: Scaling human-centered machine translation,'' \emph{arXiv preprint arXiv:2207.04672}, 2022.

\bibitem{pratap2020mls}
V.~Pratap, Q.~Xu, A.~Sriram, G.~Synnaeve, and R.~Collobert, ``Mls: A large-scale multilingual dataset for speech research,'' \emph{arXiv preprint arXiv:2012.03411}, 2020.

\bibitem{ardila2019common}
R.~Ardila, M.~Branson, K.~Davis, M.~Henretty, M.~Kohler, J.~Meyer, R.~Morais, L.~Saunders, F.~M. Tyers, and G.~Weber, ``Common voice: A massively-multilingual speech corpus,'' \emph{arXiv preprint arXiv:1912.06670}, 2019.

\bibitem{wang2020covost}
C.~Wang, A.~Wu, and J.~Pino, ``Covost 2 and massively multilingual speech-to-text translation,'' \emph{arXiv preprint arXiv:2007.10310}, 2020.

\bibitem{gales2014speech}
M.~J. Gales, K.~M. Knill, A.~Ragni, and S.~P. Rath, ``Speech recognition and keyword spotting for low-resource languages: Babel project research at cued,'' in \emph{Fourth International workshop on spoken language technologies for under-resourced languages (SLTU-2014)}.\hskip 1em plus 0.5em minus 0.4em\relax International Speech Communication Association (ISCA), 2014, pp. 16--23.

\bibitem{valk2021voxlingua107}
J.~Valk and T.~Alum{\"a}e, ``Voxlingua107: a dataset for spoken language recognition,'' in \emph{2021 IEEE Spoken Language Technology Workshop (SLT)}.\hskip 1em plus 0.5em minus 0.4em\relax IEEE, 2021, pp. 652--658.

\bibitem{jia2022cvss}
Y.~Jia, M.~T. Ramanovich, Q.~Wang, and H.~Zen, ``Cvss corpus and massively multilingual speech-to-speech translation,'' \emph{arXiv preprint arXiv:2201.03713}, 2022.

\bibitem{wang2021voxpopuli}
C.~Wang, M.~Riviere, A.~Lee, A.~Wu, C.~Talnikar, D.~Haziza, M.~Williamson, J.~Pino, and E.~Dupoux, ``Voxpopuli: A large-scale multilingual speech corpus for representation learning, semi-supervised learning and interpretation,'' \emph{arXiv preprint arXiv:2101.00390}, 2021.

\bibitem{boito2019mass}
M.~Z. Boito, W.~N. Havard, M.~Garnerin, {\'E}.~L. Ferrand, and L.~Besacier, ``Mass: A large and clean multilingual corpus of sentence-aligned spoken utterances extracted from the bible,'' \emph{arXiv preprint arXiv:1907.12895}, 2019.

\bibitem{di2019must}
M.~A. Di~Gangi, R.~Cattoni, L.~Bentivogli, M.~Negri, and M.~Turchi, ``Must-c: a multilingual speech translation corpus,'' in \emph{Proceedings of the 2019 Conference of the North American Chapter of the Association for Computational Linguistics: Human Language Technologies, Volume 1 (Long and Short Papers)}.\hskip 1em plus 0.5em minus 0.4em\relax Association for Computational Linguistics, 2019, pp. 2012--2017.

\bibitem{duquenne2022speechmatrix}
P.-A. Duquenne, H.~Gong, N.~Dong, J.~Du, A.~Lee, V.~Goswani, C.~Wang, J.~Pino, B.~Sagot, and H.~Schwenk, ``Speechmatrix: A large-scale mined corpus of multilingual speech-to-speech translations,'' \emph{arXiv preprint arXiv:2211.04508}, 2022.

\bibitem{conneau2022fleurs}
A.~Conneau, M.~Ma, S.~Khanuja, Y.~Zhang, V.~Axelrod, S.~Dalmia, J.~Riesa, C.~Rivera, and A.~Bapna, ``Fleurs: Few-shot learning evaluation of universal representations of speech,'' \emph{arXiv preprint arXiv:2205.12446}, 2022.

\bibitem{javed2022indicsuperb}
T.~Javed, K.~S. Bhogale, A.~Raman, A.~Kunchukuttan, P.~Kumar, and M.~M. Khapra, ``Indicsuperb: A speech processing universal performance benchmark for indian languages,'' \emph{arXiv preprint arXiv:2208.11761}, 2022.

\bibitem{panayotov2015librispeech}
V.~Panayotov, G.~Chen, D.~Povey, and S.~Khudanpur, ``Librispeech: an asr corpus based on public domain audio books,'' in \emph{2015 IEEE international conference on acoustics, speech and signal processing (ICASSP)}.\hskip 1em plus 0.5em minus 0.4em\relax IEEE, 2015, pp. 5206--5210.

\bibitem{radford2022robust}
A.~Radford, J.~W. Kim, T.~Xu, G.~Brockman, C.~McLeavey, and I.~Sutskever, ``Robust speech recognition via large-scale weak supervision,'' \emph{arXiv preprint arXiv:2212.04356}, 2022.

\bibitem{papineni2002bleu}
K.~Papineni, S.~Roukos, T.~Ward, and W.-J. Zhu, ``Bleu: a method for automatic evaluation of machine translation,'' in \emph{Proceedings of the 40th annual meeting of the Association for Computational Linguistics}, 2002, pp. 311--318.

\bibitem{post2018call}
M.~Post, ``A call for clarity in reporting bleu scores,'' \emph{arXiv preprint arXiv:1804.08771}, 2018.

\end{thebibliography}
\bibliographystyle{IEEEtran}

\vspace{2cm}

\section*{Authors}
\noindent {\bf Prabhjot Kaur} received an M.Sc. in Electrical Engineering from the University of Michigan-Dearborn. She is a final-year Ph.D. candidate in Computer Science at Wayne State University under the guidance of Dr. Weisong Shi. Her research spans signal processing, audio machine learning, natural language processing, and robotics.\\

\noindent {\bf Dr. L. Andrew M. Bush} is a researcher and mentor at the Utah State University Direct Lab where
he studies quadruped robotics. He has a PhD in Autonomous Systems from the Massachusetts Institute of Technology where he studied decision making under uncertainty. He has twenty-five patents due to his autonomous vehicle research at General Motors Research \& Development. He was a researcher  at MIT Lincoln Laboratory. Andy has been a visiting researcher at the NASA Ames Research Center and the Monterey Bay Aquarium Research Institute, where he developed ideas on the nature of uncertainty in reinforcement learning policies as well as designed and deployed a bathymetric mapping algorithm on an autonomous underwater vehicle.\\

\noindent {\bf Dr. Weisong Shi} is a Professor and Chair of the Department of Computer and Information Sciences at the University of Delaware (UD), where he leads the Connected and Autonomous Research (CAR) Laboratory. He is an internationally renowned expert in edge computing, autonomous driving, and connected health. Before joining UD, he was a professor at Wayne State University (2002-2022). He served in multiple administrative roles, including Associate Dean for Research and Graduate Studies at the College of Engineering and Interim Chair of the Computer Science Department. Dr. Shi also served as a National Science Foundation (NSF) program director (2013-2015) and chair of two technical committees of the Institute of Electrical and Electronics Engineers (IEEE) Computer Society. He is a fellow of IEEE, a distinguished scientist of ACM, and a member of the NSF CISE Advisory Committee.\\

\end{document}